\documentclass{Interspeech}

\interspeechcameraready

\title{In-context Language Learning for Endangered Languages in Speech Recognition}

\author[affiliation={}]{Zhaolin}{Li}
\author[affiliation={}]{Jan}{Niehues}

\affiliation[nocounter]{}{Karlsruhe Institute of Technology}{Germany}
\email{firstname.lastname@kit.edu}
\keywords{low-resource languages, large language models, speech recognition}

\usepackage{comment}

\begin{document}

\maketitle

\begin{abstract}
    
    With approximately 7,000 languages spoken worldwide, current large language models (LLMs) support only a small subset. Prior research indicates LLMs can learn new languages for certain tasks without supervised data. We extend this investigation to speech recognition, investigating whether LLMs can learn unseen, low-resource languages through in-context learning (ICL). With experiments on four diverse endangered languages that LLMs have not been trained on, we find that providing more relevant text samples enhances performance in both language modelling and Automatic Speech Recognition (ASR) tasks. Furthermore, we show that the probability-based approach outperforms the traditional instruction-based approach in language learning. Lastly, we show ICL enables LLMs to achieve ASR performance that is comparable to or even surpasses dedicated language models trained specifically for these languages, while preserving the original capabilities of the LLMs. Our code is publicly available \footnote{\url{https://github.com/ZL-KA/ICLL}}.
\end{abstract}

\section{Introduction}
State-of-the-art automatic speech recognition (ASR) systems have achieved human-level performance for high-resource languages. However, their performance remains limited for endangered languages \cite{le-ferrand-etal-2024-modern, chen-etal-2024-towards-robust, pratap2024scaling}. Popular ASR systems incorporate both acoustic and language information. Previous studies have shown that the effectiveness of leveraging acoustic information gets significantly improved for endangered languages, but the integration of language information remains under-explored \cite{liu-etal-2024-important, li-niehues-2025-enhance}.

With the rise of large multilingual language models (LLMs), many such models are now available. 

Nevertheless, their application in automatic speech recognition (ASR) systems for endangered languages, especially those that are not represented in the training data, remains unclear. In addition to traditional fine-tuning methods, LLMs offer an alternative approach to learning new tasks known as In-Context Learning (ICL). This enables LLMs to perform tasks by conditioning on a few example prompts, without requiring parameter updates \cite{brown2020language}. Instead of fine-tuning, ICL leverages the model’s pre-trained knowledge and adapts it to new tasks.

While ICL has demonstrated strong performance for high-resource languages with abundant training data, its effectiveness drops significantly for low-resource languages \cite{aji-etal-2022-one, asai-etal-2024-buffet}. This decline is primarily due to insufficient linguistic representation in the model’s pretraining data, leading to poor generalization and making ICL less reliable for these languages.

To enhance LLM performance on low-resource languages, one research direction is cross-lingual ICL (X-ICL) \cite{brown2020language, winata2022cross, cahyawijaya-etal-2024-llms}.  X-ICL leverages the model’s multilingual knowledge by providing prompts in a high-resource language or a mix of high- and low-resource languages while expecting the output in the low-resource language. This approach facilitates knowledge transfer across linguistic boundaries and mitigates data scarcity. However, X-ICL often introduces translation biases and struggles with structural inconsistencies between languages. Additionally, its performance tends to decline in extremely low-resource scenarios, particularly for languages the model has never encountered.

Another research direction is long-context ICL, which extends the context length available for in-context learning, enabling models to process more examples and capture richer patterns \cite{bertsch-etal-2025-context, ginn-etal-2024-teach}. By incorporating longer prompts with more diversity, long-context ICL can improve generalization to low-resource languages. However, it comes with challenges such as increased computational cost, sensitivity to input ordering, and diminishing returns when context length exceeds the model’s optimal processing capacity.

\begin{table*}[h]
\centering
\begin{tabular}{cccccccc}
\hline
Language  & ISO code & Language Family  & Audio source                        & Train samples  & Train (h) & Dev+Test (h) \\ \hline
Khinalug & kjj & Northeast Caucasian & Spontaneous                       & 978    & 2.14        & 0.49      \\
Kichwa   & que &  Quechuan            & Radio                          & 2991       & 3.05        & 0.77    \\
Mboshi   & mdw &  Bantu ZoneC         & Reading                        & 4513       & 3.93        & 0.53    \\
Japhug   & jya & Sino-Tibetan        & Spontaneous                    & 23975       & 27.74       & 7.00     \\   \hline
\end{tabular}
\caption{Dataset descriptive statistic}
\label{tab:dataset}
\end{table*}


Beyond exploring approaches to boost ICL performance on low-resource languages, researchers have investigated different evaluation strategies to assess ICL performance across various tasks. Common evaluation benchmarks for low-resource languages include machine translation \cite{zhu-etal-2025-evaluating}, which tests cross-lingual transferability; code-switching \cite{zhang-etal-2023-multilingual}, which evaluates the model’s ability to handle mixed-language inputs; interlinear text glossing \cite{ginn-etal-2024-teach}, which examines morphological and syntactic understanding. However, the Automatic Speech Recognition (ASR) task remains significantly underexplored. Unlike text-based tasks, ASR faces unique challenges, such as the acoustic variability caused by diverse accents and speaker demographics, the complex phonetic and prosodic features that are often poorly represented in pretrained models. Moreover, low-resource languages sometimes lack standardized orthography, making ASR even more difficult compared to purely tasks-based tasks.

Despite these challenges, some studies have attempted to apply ICL in ASR for low-resource languages. \cite{hsu2024meta} explored ICL by feeding relevant audio-text pairs alongside the target audio into ASR models. However, long audio inputs restrict the number of ICL samples that can be used, limiting overall performance. Additionally, their approach required LLMs that support audio input, further narrowing the choice of models.


Motivated by previous research, we propose not only learning a new task through ICL but also a new language. By In-context language learning (ICLL),  LLMs can leverage their knowledge of multiple languages to learn an entirely new language using just a few hundred sample sentences. We evaluate this capability in the context of ASR, assessing how well LLMs can adapt to new languages for this task. Additionally, ICLL offers the advantage of selecting different samples for each test instance, allowing better adaptation to individual text examples. To optimize performance, we investigate various sample selection strategies.

We summarize our main contributions below.
\begin{itemize}
    \item ICLL enables LLMs to learn a new language with only a few hundred samples.
    \item ICLL achieves better or comparable performance to corpus-trained language models in language modelling and ASR tasks.
    \item Sample selection in ICLL is important to achieve strong performance.
\end{itemize}

\section{In-context Language Learning} \label{sec:ICL}

This work aims to investigate language modelling with ICL for low-resource languages, which means the objective is not to teach a new task to LLMs through examples within the prompt, but to provide linguistic knowledge of the target language to LLMs. Even though, the general setup for both objectives remains the same.

The setup begins with prompt creation, which consists of sentences in the target language without additional instructions. Due to computational resources constraints, it is impractical to feed all available text within the prompt, even in the scenarios of low-resource languages. To address this, we investigate various sample selection strategies with explanations in Section \ref{sec:ICL_retrieval}.

Once the prompt is provided, LLM generates text in the target language with task-specific requirements to perform ICL. However, in the use case of ASR, this work focuses on not generating text but on ranking different possible transcriptions. To achieve this, we use an initial ASR system to generate an n-best list of candidate transcriptions, which is subsequently re-ranked using ICL. We explore different methods for selecting the best matching hypothesis to optimize this process, as explained in Section \ref{sec:ICL_selection}.

\subsection{Samples Retrieval Strategy} \label{sec:ICL_retrieval}

The baseline approach is to select random samples. As the most general method for learning the language, this may not lead to the best performance since the samples do not align with the specific task \cite{ginn-etal-2024-teach, hsu2024meta}.

In traditional language modelling, \textbf{corpus-level selection} is typically performed, where a single set of samples is selected for the test set. This is because training a language model requires substantial computation. However, the drawback is that different test instances may require different samples, which corpus-level selection does not account for. In ICL, we have the advantage that no additional training for the language model is needed. Therefore, we can select different ICL samples for each test example (\textbf{example-specific selection}).

In addition to when to select the samples, we also investigate how to select the samples. We use the training examples as our search database and treat each test example as a query. We then select the most similar training examples based on the cosine similarity of embeddings computed by an additional embedding model. Depending on the experiment, we vary the choice of key used for searching with different strategies.

As a first approach, we use the best hypothesis from the initial ASR system as the query (\textbf{hyp}). In corpus-level selection, we compute the average similarity scores across all test examples for each training example to obtain a single ranking of the training sentences.

Since this might bias the model towards the best initial hypothesis,  we also use the top $k$ hypothesis (\textbf{topK}). In our experiments, we used $k=3$. In the sample selection, we then average the $k$ similarities for each training example. For the corpus-level selection, we apply the same approach but across the entire corpus. The final ranking of training samples is determined by summing up their similarity scores within the corpus.

Finally, we also investigate whether we can directly use the audio sample. Since we use a multi-modal embedding model, we can compare embeddings of different modalities. Specifically, we compare the embeddings of test audio examples with those of training audio examples (\textbf{audio}). Additionally, we explore combining audio and text embeddings by summing them.  We limit these approaches to the sample-specific due to their inferior preliminary results compared to other methods. 

To evaluate how good the sample retrieval is, we conduct an \textbf{oracle} experiment, where the human transcript is used as the query.

\begin{table*}[h]
\centering
\begin{tabular}{cccccccccccccc}
\hline
 & Level & Strategy & 0 & 1 & 5 & 10 & 20 & 50 & 100 & 150 & 200 & 250 & 300 \\ 
 
\hline
Khinalug & Corpus &  random & 1142 & 228 & 148 & 114 & 99 & 80 & 57 & OOM &  &  &  \\ 
 & Corpus & hyp & 1142 & 331 & 191 & 200 & 161 & 145 & 86 & 70 & 62 & OOM &  \\ 
 & Corpus & topk & 1142 & 331 & 191 & 182 & 156 & 122 & 86 & 66 & 64 & OOM &  \\ 
 & Example & oracle & 1142 & 181 & 59 & 47 & 37 & 31 & 29 & OOM &  &  &  \\ 
 & Example & hyp & 1142 & 217 & 76 & 56 & 44 & 36 & OOM &  &  &  &  \\ 
 & Example & topk & 1142 & 208 & 71 & 56 & 45 & 36 & OOM &  &  &  &  \\ 
 & Example & audio & 1142 & 303 & 158 & 119 & 86 & 70 & OOM &  &  &  &  \\ 
 & Example & hyp\&audio & 1142 & 220 & 95 & 62 & 51 & 44 & OOM &  &  &  &  \\ 
 \cline{2-14}
 & \multicolumn{2}{l}{n-gram LM same vocab} & \multicolumn{11}{c}{133} \\
 & \multicolumn{2}{l}{Trans LM same vocab} & \multicolumn{11}{c}{49} \\
\hline
Kichwa & Corpus & random & 6978 & 1076 & 221 & 251 & 211 & 143 & 132 & 106 & 146 & 85 & OOM \\ 
 & Corpus & hyp & 6978 & 2221 & 1244 & 1058 & 642 & 645 & 308 & 270 & 275 & 226 & 186 \\ 
 & Corpus & topk & 6978 & 2221 & 1392 & 508 & 486 & 362 & 323 & 269 & 214 & 189 & 150 \\ 
 & Example & oracle & 6978 & 359 & 85 & 53 & 42 & 37 & 31 & 44 & OOM &  &  \\ 
 & Example & hyp & 6978 & 366 & 101 & 76 & 46 & 37 & 31 & 41 & 31 & OOM &  \\ 
 & Example & topk & 6978 & 375 & 102 & 71 & 43 & 37 & 32 & 41 & OOM &  &  \\ 
 & Example & audio & 6978 & 2216 & 178 & 113 & 96 & 74 & 71 & 72 & OOM &  &  \\ 
 & Example & hyp\&audio & 6978 & 3693 & 193 & 114 & 91 & 63 & 43 & 46 & OOM &  &  \\ 
 \cline{2-14}
 & \multicolumn{2}{l}{n-gram LM same vocab} & \multicolumn{11}{c}{78} \\
 & \multicolumn{2}{l}{Trans LM same vocab} & \multicolumn{11}{c}{35} \\
\hline
\end{tabular}
\caption{Perplexity evaluation for different samples retrieval Strategies on Khinalug and Kichwa datasets. The columns labeled with numbers represent the number of samples in ICL. The selection is implemented with a single RTX6000 GPU with 48 GB of memory.}
\label{tab:sample_selection_ppl}
\end{table*}

\subsection{ASR Hypotheses Re-ranking} \label{sec:ICL_selection}

The task of hypothesis selection is to select one hypothesis from the $n$-best lists generated by the initial ASR system. The traditional LLM-based approach to hypothesis selection is instruction-based, where the LLM is explicitly instructed to choose one from the provided options given the ICL samples; However, since the LLM is not instruction-tuned for the target language, we hypothesize that the effectiveness of this method may be limited. Furthermore, the instruction-based approach cannot easily integrate the knowledge from the acoustic model.

Therefore, we also investigate a second approach to select the best hypothesis based on the language model probability assigned to the hypothesis. At first, we solely rely on the language model probability, assuming that all the hypotheses in the n-best list have a high acoustic probability. Then, we combine the language model and acoustic probability according to the following equation:

\begin{equation}
    S = \sum_{i=1}^{N} \log P_{\text{AM}}(x_i) + \sum_{j=1}^{M} \log P_{\text{LM}}(y_j)
\end{equation}

where $P_{\text{AM}}$ represents the probability from the acoustic model at time step $i$ with a total of $N$ steps, and $P_{\text{LM}}$ represents the probability from the language model at time step $j$ with a total of $M$ steps.

\begin{table}[h]
\centering
    \begin{tabular}{ccccc} \hline
    & Instruct-10 & Instruct-50 & CE &  CE + AP  \\  \hline
    Khianlug & 45.16 & 65.44 & 42.30 & 42.12  \\
    Kichwa & 33.54 &  80.59 & 17.05 & 17.15  \\
    Mboshi & 38.42 & 80.59 & 34.11 & 30.64 \\
    Japhug & 26.36 & 40.16 & 23.90 & 23.86  \\ 
    \hline
    \end{tabular}
    \caption{Hypotheses selection results with instruction-based and probability-based approaches. CE stands for cross-entropy and AP stands for acoustic probability; Instruct-10 indicates having 10 samples in ICL; The number of samples in ICL for Khinalug and Japhug is 50, and that for Kichwa and Mboshi is 150. The settings are based on perplexity evaluation as referred to Table \ref{tab:sample_selection_ppl} under computational resource limitation.}
\label{tab:hypotheses_selection_asr}
\end{table}

\begin{table*}[h]
    \centering
    \begin{tabular}{cccccccc} 
    \hline

    & Acoustic & n-gram LM & Trans LM & ICL (hyp) & \textbf{ICL (topk)} & ICL (oracle)  & Oracle  \\ \hline
    Khinalug & 42.12 & 39.63 & 41.57 & 42.12 & 41.84 & 41.66 & 36.50 \\
    Kichwa  & 17.31 & 17.72 & 18.85 & 17.15 & 17.00 & 16.90 & 12.42 \\
    Mboshi  & 31.14 & 30.64 & 30.87 & 30.64 & 30.24 & 30.10 & 22.72 \\ 
    Japhug & 23.94 & 23.29  & 23.34 & 23.86  & 23.81 & 23.86 & 20.07  \\ 
    \hline
    \end{tabular}
    \caption{ICL results on ASR hypotheses selection. }
\label{tab:asr_systems}
\end{table*}
\section{Experimental Setups}

\subsection{Datasets}

To address the unique challenges for low-resource languages, such as language complexity, limited corpus size, and sparse audio sources, this study conducts experiments on four linguistically diverse languages: Khinalug \cite{li-etal-2024-speech}, Kichwa \cite{taguchi-etal-2024-killkan}, Mboshi \cite{godard-etal-2018-low}, Japhug \cite{guillaume-etal-2022-fine}. All selected languages are recognized as endangered and unseen to LLMs. The description and statistical overview of the datasets is available in Table \ref{tab:dataset}.

\subsection{Hypotheses Generation}

We utilize the state-of-the-art version of Wav2Vec2 model mms-300 \footnote{\url{https://huggingface.co/facebook/mms-300m}} to generate ASR hypotheses. Pre-trained with over 1400 languages, the model provides extensive linguistic coverage and adaptability for low-resource settings \cite{pratap2024scaling}. In decoding, we implement beam search, and the 10 hypotheses with the best acoustic probabilities are chosen for language model selection.

\subsection{Language Models}
In this work, we employ Llama-3-8B \footnote{\url{https://huggingface.co/meta-llama/Meta-Llama-3-8B-Instruct}} as the large language model due to its success in language learning. We utilize 5-gram LMs as statistic LM and GPT-2 models as transformer-based LMs. 



\subsection{Embedding Similarity Measurement}

For ICL sample selection, we use SONAR \cite{Duquenne:2023:sonar_arxiv} to generate embeddings for both hypotheses and audio. As a state-of-the-art multilingual representation learning model, SONAR supports a wide range of languages, making it especially useful for embedding the low-resource languages in this study, which are not supported by all models.

\section{Results and Analysis}

\subsection{Language Modelling} \label{s4_1}

In the first step, we evaluated how good language learning with ICL performs on direct language modelling. We provide LLMs both the retrieved samples and the texts of the target samples, and calculate the perplexity only for the texts. In this case, we measure the performance in perplexity and show the results in Table \ref{tab:sample_selection_ppl}.

As baselines, we additionally train an n-gram language model and a transformer-based language model using the same text encoding as the LLM. Since all experiments share the same vocabulary, comparing their results is valid and meaningful.

\textbf{Corpus-level retrieval works}: As seen in the first two rows of the table, corpus-level retrieval achieves performance similar to selection using n-gram LMs for each language but falls short compared to transformer-based LMs. This indicates that this strategy serves as a strong baseline for comparison with other retrieval strategies.

\textbf{Example-specific retrieval benefits}: As shown in the following rows of Table \ref{tab:sample_selection_ppl}, selecting example-specific ICL samples leads to better performance than corpus-level selection. Furthermore, there is little difference between using one or three ASR hypotheses as search keys. Although we initially expected that multiple hypotheses would introduce greater linguistic diversity and lead to better sample selection, all hypotheses originate from the same speech and ASR models, limiting their actual diversity.

\textbf{Audio similarity harms}: We explore retrieval based on acoustic similarity and observe that using either acoustic embeddings alone or a combination of acoustic and text embeddings results in lower performance compared to text embeddings alone. This suggests that acoustic embeddings do not significantly contribute to retrieving high-quality ICL samples. A possible explanation is that these endangered languages are not closely related to those supported by the embedding model, leading to ineffective retrieval.

\textbf{ASR errors for retrieval less important}: we observe that using a single ASR hypothesis as the search key yields results comparable to the oracle experiment, which uses the gold transcript. This holds despite the high word error rate of the ASR models, suggesting that retrieval remains effective even in the presence of ASR errors.

\subsection{LLM-based Hypotheses Selection} \label{sec:hypo_selection_results}

This work explores instruction-based and probability-based approaches for ASR hypothesis selection. Specifically, we examine cross-entropy along with the impact of incorporating acoustic probabilities from acoustic modelling. Performance is evaluated using Word Error Rate (WER) for ASR tasks.

Table \ref{tab:hypotheses_selection_asr} shows that the instruction-based approach performs poorly, likely because the LLM struggles to learn linguistic knowledge using only ICL samples. Besides, we find that increasing the ICL samples from 10 to 50 leads to clear performance degradation, and we suppose the linguistic diversity within ICL samples brings a negative impact on the instruction-based approach. 

In contrast, probability-based approaches achieve significantly better results. Additionally, incorporating acoustic information consistently improves hypothesis selection for all methods. Besides, the linguistic diversity benefits to probability-based approach since more samples leading to better perplexity scores (Table \ref{tab:sample_selection_ppl}).

\subsection{ICL in Speech recognition}

After understanding the capabilities of LLMs in ICLL for speech recognition, we evaluate how well they perform compared to other approaches. As shown in Table \ref{tab:asr_systems}, we experiment with several selection methods: \textit{Acoustic}, which selects the hypothesis with the highest acoustic probability; \textit{N-gram LM} and \textit{Trans LM}, which select the hypothesis with lowest perplexity score with the evaluation of the LMs; \textit{WER Oracle}, which selects the best hypothesis according to the lowest WER among the hypotheses to the ground-truth; and \textit{ICL(hyp), ICL(topk), ICL(oracle)}, which represents our approach of with hyp strategy at corpus level, topk and oracle strategies at example level. The acoustic probability is incorporated into ICL experiments, motivated by the results in Section \ref{s4_1}.

As shown in Table \ref{tab:asr_systems}, all ICL-based selection methods outperform the acoustic selection across all languages, demonstrating that LLMs effectively learn the target languages when provided with ICL samples. Additionally, ICL selection achieves better performance than n-gram and transformer-based LMs for Kichwa and Mboshi and comparable performance for the remaining two languages, despite using significantly fewer samples in ICL than LMs used in training, while preserving the LLM’s original capabilities. Notably, ICL selection is implemented using limited computational resources (a single 48GB RTX6000 GPU). Given that perplexity continues to decrease as more samples are used (as shown in Table \ref{tab:sample_selection_ppl}), we assume ICL selection has the potential to achieve even better results with greater computational resources.

\subsection{Limitations}

One limitation of this work is that probability-based selection requires access to model parameters, which restricts its application to open-source LLMs. Additionally, our experiments focus on four low-resource languages from different language families and with varying resource availability. While our findings are consistent across these languages, their applicability to all other languages remains uncertain. Moreover, we evaluate our approach using only one LLM, Llama v3. Although it is among the most powerful models, other LLMs may have even stronger multilingual capabilities. We assume our approach can be applied to these models as well, but its actual performance across different LLMs remains unknown.

\section{Conclusion}

In this work, we propose In-context language learning (ICLL) that enables large language models to learn a new language through in-context examples. We show that sample selection strategies for ICLL are essential for achieving strong performance. This is also highlighted by the gains of example-specific selection compared to corpus-level selection. However, the hypothesis of an initial system is a good proxy for selection, leading to nearly the same performance as the oracle human transcriptions. We evaluated the approach on 4 endangered languages and showed better performance in language modelling measured in perplexity compared to strong baselines. Furthermore, when integrated into an ASR system as a re-ranking step, we could also improve the performance in two out of four languages.

\section{Acknowledgements}

This work is supported by the Deutsche Forschungsgemeinschaft (DFG) under the project Computational Language Documentation by 2025 (CLD 2025). We acknowledge the HoreKa supercomputer funded by the Ministry of Science, Research and the Arts Baden-Wurttemberg and by the Federal Ministry of Education and Research.

\bibliographystyle{IEEEtran}
\bibliography{mybib}

\end{document}